\documentclass[lettersize,journal]{IEEEtran}
\usepackage{amsmath} % assumes amsmath package installed
\usepackage{graphicx}
\usepackage{makecell}
\usepackage{amsfonts}
\usepackage{amssymb,bm}
\usepackage{import}
\usepackage{xifthen}
\usepackage{pdfpages}
\usepackage{transparent}
\usepackage{subcaption}
\usepackage{hyperref}

\usepackage[ruled,vlined,linesnumbered]{algorithm2e}
\usepackage{algpseudocode}
\usepackage{float}
\usepackage{multirow}
\usepackage{booktabs}

 \usepackage{soul} 
 \usepackage{array,colortbl,xcolor}
\newcommand{\highlighti}[2]{%
    \setlength{\fboxrule}{1pt}%
    \fcolorbox{#1}{lightgray}{#2}}

\newcommand \figref[1]{Fig. \ref{#1}}

\newcommand\StateSpace[1]{\mathbf{\mathcal #1}}
\newcommand\CS{$\mathcal{C}$}
\newcommand\CF{$\mathcal{C}_{free}$}
\newcommand\NaturalNumbers{\mathbb{N}}

\newcommand\MinusSet{\backslash}
\newcommand\Real{\mathbb{R}}
\renewcommand\vector[1]{\bm{#1}}
\newcommand\ExpectedValue{\mathbb{E}}
\newcommand\BehavesLike{\sim}
\newcommand\Norm[1]{\|#1 \|_2}
\newcommand\Gradient{\nabla}
\newcommand\Normal{\mathcal{N}}

\newcommand{\cfree}{$C_{free}$}

\title{\LARGE \bf
Enhancing Path Planning Performance through Image Representation Learning of High-Dimensional Configuration Spaces}

\author{Jorge Ocampo Jimenez and Wael Suleiman% <-this % stops a space
\thanks{This work was partly supported by Natural Sciences and Engineering Research Council of Canada (NSERC). J. Ocampo Jimenez is a doctorate student funded by Consejo Nacional de Ciencia y Tecnología (CONACyT, Mexico City, Grant No. 278823) and by the merit scholarship programme for foreign students (PBEEE) by the Fonds de recherche du Québec – Nature et technologies.}% <-this % stops a space
\thanks{Authors are with Electrical and Computer Engineering Department, Universit\'e de Sherbrooke, Quebec, Canada  (e-mail: \{Jorge.Ocampo-Jimenez; Wael.Suleiman\}@USherbrooke.ca)}%

}

\begin{document}
\maketitle
\thispagestyle{empty}
\pagestyle{empty}

\begin{abstract}
This paper presents a novel method for accelerating path-planning tasks in unknown scenes with obstacles by utilizing Wasserstein Generative Adversarial Networks (WGANs) with Gradient Penalty (GP) to approximate the distribution of waypoints for a collision-free path using the Rapidly-exploring Random Tree algorithm. Our approach involves conditioning the WGAN-GP with a forward diffusion process in a continuous latent space to handle multimodal datasets effectively.
We also propose encoding the waypoints of a collision-free path as a matrix, where the multidimensional ordering of the waypoints is naturally preserved. This method not only improves model learning but also enhances training convergence.
Furthermore, we propose a method to assess whether the trained model fails to accurately capture the true waypoints. In such cases, we revert to uniform sampling to ensure the algorithm's probabilistic completeness; a process that traditionally involves manually determining an optimal ratio for each scenario in other machine learning-based methods.
Our experiments demonstrate promising results in accelerating path-planning tasks under critical time constraints. The source code is openly available at https://bitbucket.org/joro3001/imagewgangpplanning/src/master/.
\end{abstract}


\section{Introduction}

% \begin{figure}[h!]
% \centering
% \includegraphics[width=0.6\linewidth]{training2dProjection}
%     \caption{Image representation of a collision-free shortest path for a 7D manipulator robot. The original shortest path waypoints are clustered, with the state values scaled between [0,1] and encoded as a matrix based on the current state of the workspace. This data is used to train an image-to-image model, which takes the workspace as an RGB-D image input and generates a matrix representation of the path waypoints. Our proposed model, using WGAN-GP, effectively approximates the constrained \CF{}. The ordering of the waypoints follows a continuous transition between states, contributing to stable training.}
%
% \end{figure}

Machine learning techniques, such as neural network generative models like Variational Autoencoders (VAEs) \cite{qureshi2019motion} and Generative Adversarial Networks (GANs) \cite{DBLP:journals/ral/LembonoPJC21}, have been used to improve the efficiency of random sampling algorithms. These models bias the sample distribution towards collision-free states, conditioned on the robot's current workspace configuration. While previous research has successfully applied GANs \cite{DBLP:journals/ral/LembonoPJC21,JAS-2021-0110,Li2021EfficientHG} to tasks such as generating inverse and forward kinematics, there has been limited exploration of using GANs with gradient penalties in high-dimensional problems. This is largely due to the increased complexity and computational cost associated with such approaches.

Moreover, it has been demonstrated that Wasserstein GANs (WGANs) are highly sensitive to hyperparameter tuning. This sensitivity arises because the transport cost between the image and waypoints of a collision-free path derived from a training dataset is often represented by discontinuous functions \cite{rout2022generative}.

Another challenge in employing machine learning algorithms for waypoint sampling is the inherent inaccuracy of the learned model. Since the model's predictions are not perfectly accurate, the sampling process uses a predefined ratio to balance between using the uniform distribution and the learned distribution to ensure probabilistic completeness \cite{LearningSampling}. However, determining the optimal ratio requires careful tuning, which can pose additional difficulties.

\section{Proposed method}
To address challenges such as unstable training processes, extended training times, and the potential for unexpected outcomes when encountering unseen conditions during path planning, we propose to:
\begin{itemize}
    \item Select a forward diffusion process from an SDE model to train the input condition of the workspace. Unlike using an encoder from a VAE, the SDE approach does not require joint training with the generative model, simplifying the training process while maintaining effective performance.

    \item Reduce the number of images in the input of the trained network by using affinity propagation. This process reduces the number of samples per epoch during training by representing the whole waypoint set as a matrix. 

    \item Exploit in the cases of bounded configuration spaces, the fact that the samples clustered by the affinity points are bounded too; thus,  we could reject samples too far from their means in the bounded sets; removing the need of finding an optimal sampling ratio between the uniform distribution and the learned one.

\end{itemize}
Our model is trained using forward kinematics simulations with the Baxter manipulator robot. We create multiple scenarios by placing random human models around the robot and using RGB-D representations of the robot's obstacles to train our model to estimate the waypoints of a collision-free path. 

We evaluate the performance of our model using metrics such as planning time, path length, and success rate. This evaluation enables us to assess the impact of changes in the conditions and encoding of configuration states on the model's ability to generate new waypoints. To establish a baseline for comparison, we employed paths generated by RRT and RRT*. Our experiments demonstrated that our model is capable of finding shorter and faster paths in time-constrained scenarios. %We believe that the proposed model holds substantial potential for tasks requiring rapid, collision-free sampling. This is particularly advantageous when an RGB-D representation of the robot's surroundings is available.

\section{Original Contributions}
This paper offers several novel contributions, including:
\begin{enumerate}
    \item Introducing a new architecture that enhances WGAN-GP training by embedding the condition as an additional channel in the RGB-D representation of the working space (WS).
    \item Reducing the number of queries by generating all the waypoints in one query.
    \item Introducing a new methodology that discards samples from the trained model during deployment of the planner when a probability threshold is met in bounded configuration spaces, thus preserving probabilistic completeness.
    \item Enhancing path planning performance in terms of time, success rate, and path length compared to algorithms like RRT and RRT* under time constraints.

\end{enumerate}
Compared to other similar works where the planner's sampling is biased, our method offers several advantages, including:
\begin{itemize}
    \item By utilizing both GANs and forward diffusion, our model can handle noisy or previously untrained scenarios while generating high-quality WGAN samples.
    \item Our image-to-image model has the potential to extend to the prediction of higher-dimensional waypoints for robotic tasks using image processing algorithms, thereby reducing the number of connections between networks by employing pattern-based deep neural network models.
    \item Unlike other machine-learning methods \cite{qureshi2019motion,LearningSampling} that depend on finding an appropriate ratio between uniform or biased distributions for RRT-based planners, our approach is established an error bound to reject the sampling of the trained model.
\end{itemize}

\begin{figure*}[!ht]
\begin{center}
\includegraphics[width=0.83\linewidth]{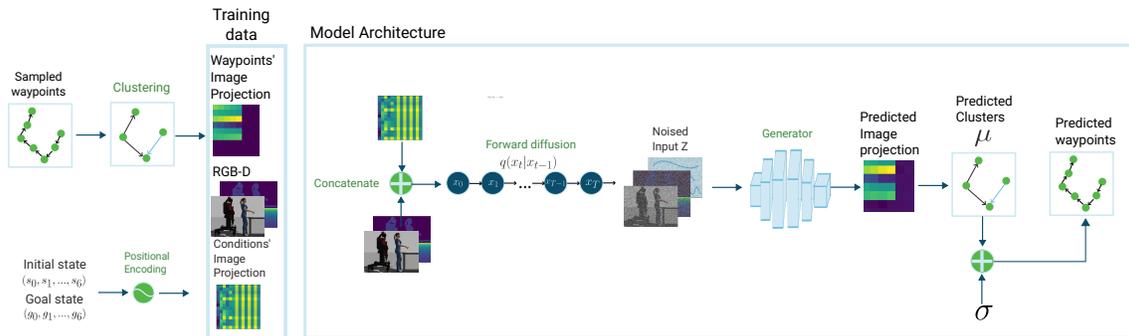}
    \caption{Proposed architecture to learn the waypoints of a collision-free path from the robot's WS.}
\label{fig:model}
\end{center}

\end{figure*}

\section{Related Work}
The use of learning by demonstration has proven to be an effective approach in numerous studies aimed at enhancing the performance of sampling-based random planners \cite{9712347}. A common technique involves employing an auto-generative model that learns a mapping between a robot's configuration space (\CS{}) and the image-based scenario, using a reduced number of samples from the full distribution. In recent years, deep neural network (DNN) models have gained significant traction in this field due to their ability to process large volumes of input data, such as image or point cloud representations of the robot's environment, and to generalize across a wide range of examples, including potential robot configurations and the number and location of obstacles in the workspace.

While DNNs allow conditioning in high-dimensional spaces; they are trained as unimodal models, which fail to capture the inherent diversity and multiple modes in the data. In contrast, Deep Generative models have demonstrated their ability of capturing high-dimensional multimodal data in contexts like text and image generation \cite{Urain:2024vn}.

Generative models are widely used in the context of Rapidly-exploring Random Trees-based algorithms \cite{Kuffner} for two main purposes: to introduce a bias in the sampling process or to serve as a heuristic for the cost function. These models guide the algorithm towards lower-cost paths by taking into account the specific conditions of the scenario.

The application of neural networks for learning the sampling distributions to bias-sampled based planners was first introduced in \cite{LearningSampling}. The study utilized a conditional variational autoencoder to identify areas in \CS{} that held promise based on the initial and goal states, as well as the obstacles present in the scenario. This enabled the sampler of random path algorithms to be biased, resulting in more efficient path planning.

In another study \cite{qureshi2019motion}, an encoder was used to capture environmental information, with the sampler conditioned on raw sensor data or voxelized output embedded in the latent space. The encoded information was then utilized by a planning network, in conjunction with the current and goal states, to generate the next state. This model can bias the sampler of RRT* \cite{Karaman} and has been tested on high-dimensional configuration spaces.

The research presented in \cite{JAS-2021-0110} utilizes 2D working spaces as inputs for a conditional GAN. The GAN is conditioned on the RGB representation of both the initial and final points of the path, as well as the map of the working space. The generator is trained with two discriminators: one for the obstacle map and another for the initial and final goal states represented in the working space. The resulting algorithm achieves an impressive success rate of approximately 90\% in generating connected configurations.

The study presented in \cite{teachingRobot} utilizes inverse reinforcement learning to determine the weights of the RRT* cost function based on the expected behavior of a robot in environments previously occupied by humans. This approach helps guide the planner towards the desired path. However, it may be less suitable for dynamic environments where the weights cannot be adjusted without compromising the asymptotic optimality of the algorithm.

The authors of \cite{Li2021EfficientHG} present an approach where GANs are used to bias an RRT-based planner by incorporating Encoders and Decoders directly as hidden layers in the generator. The initial state, map, and latent vector are provided as inputs to the encoder, while the decoders output a 2D representation of the path. The generator thus produces the path as an output image, treating the task as an image-to-image model. Although the authors do not provide specific information on running times, it is reported that the algorithm requires fewer iterations to achieve a lower cost compared to RRT*.

The research presented in \cite{DBLP:journals/ral/LembonoPJC21} explores the use of a GAN to learn the inverse kinematics of high-dimensional robots. The model is conditioned on the target working space position of the end effectors, enabling the generation of samples in high-dimensional \CS{}s, which was previously infeasible. However, it is important to note that the conditioning is not directly based on sensor data or the current state of the scenario.

Lately, significant advances in generative neural network models have been made, such as the diffusion generation approach in path planning \cite{huang2023diffusion}, their application in random sampling-based planners remains limited, particularly in scenarios with strict time constraints for obtaining a collision-free path. A common challenge with diffusion-based methods is their computational cost; repeatedly sampling from the diffusion generator during the backward pass can be time-consuming. For a more comprehensive discussion on the use of generative DNN models, we refer to the work of \cite{Urain:2024vn}.

\section{Problem Formulation}

The objective of this research is to develop a method for approximating waypoints of a collision-free path by leveraging information from its obstacles, represented as an image-scenario. The ultimate goal is to enhance the performance of a path planner. This is achieved by training a model to learn the mapping from the image-scenario to \CS{}, which allows for more efficient sampling and accelerates the planning process. The proposed method has the potential to significantly improve robotic systems' performance by reducing the computational cost of planning while still producing low cost paths.

Mathematically speaking, a path planning problem is defined by a configuration space
$\StateSpace{C}$$=[0,1]^d$
with dimension
$d \in \NaturalNumbers, d\geq 2$; a $\StateSpace{C}_{obs}$ that 
is defined as the set of \CS{} that corresponds to the collision states; and a free configuration space 
$\StateSpace{C}_{free}=\StateSpace{C} \MinusSet \StateSpace{C}_{obs}$
, with initial configuration 
$\mathit{q}_{0} \in \StateSpace{X}_{free}$
and a  set of goal configuration 
$\StateSpace{C}_{goal} \subset \StateSpace{C}_{free}$. A path is a continuous function
$s:[0,1]\rightarrow \Real^d$
, and it is collision-free if
$s(\tau)\in \StateSpace{C}_{free}$
for all
$\tau \in [0,1]$
and feasible when it is collision-free and
$s(0)=\mathit{q}_{0}$ and $s(1) \in \StateSpace{C}_{goal}$.

Finding a feasible path in the \CS{} of a robot is known to be PSPACE-complete \cite{10.5555/1213331}, which makes it computationally intractable for most practical applications. Consequently, researchers have developed sampling-based motion planning algorithms to address this challenge in high-dimensional \CS{}s. These algorithms work by randomly sampling configurations and connecting them to form a path to the goal states. Completeness, or finding a solution if one exists, requires drawing a sufficient number of uniformly distributed random samples. Asymptotic optimality, where the path cost converges to the optimal solution, can be achieved by systematically connecting the nodes of the search tree \cite{LearningSampling}.

To improve the efficiency of sampling-based motion planners, researchers have proposed various methods to bias the path towards the goal. One such method involves learning a probability distribution over the waypoints in $s(\cdot)$ based on the robot's scenario, which guides the sampling process to explore regions of \CS{} more likely to lead to the goal. This approach reduces the time spent exploring regions of \CS{} that are less likely to yield a viable path, thereby accelerating the planning process. As a result, it has the potential to substantially enhance the efficiency of sampling-based motion planning algorithms, making them more practical for real-world applications.

\section{Methodology}\label{sec:methodology}

We propose a novel approach for accelerating sampling-based motion planning algorithms by generating waypoints of a path in \(\StateSpace{C}_{free}\) with additional properties such as feasibility and connectivity with the current path. Our approach utilizes a WGAN to sample from a learned distribution over \(\StateSpace{C}_{free}\), which biases the sampling process towards regions of \CS{} more likely to yield waypoints. Specifically, we employ a WGAN-GP to generate high-quality collision-free configurations without the need to determine a suitable clipping interval. This method replaces the uniform distribution typically used for sampling \CS{} and results in faster query times. The proposed architecture is illustrated in \figref{fig:model}. In this architecture, we simplify the original number of waypoints-states by learning the centroid of clusters when the number of waypoints is higher than the matrix representation; then, the clusters follow an already multidimensional sort given the order of the waypoints in a path; this increases the chances of having smooth gradients during training. New waypoint clusters are generated by applying positional encoding to the initial and final states of the desired path to an RGB-D image representation of the current WS of the robot. The RGB-D is then diffused through a continuous stochastic process, which serves as the latent space of a generative neural network capable of predicting the sorted multidimensional waypoints. To recover any potential missing states of a collision-free path within these clusters, sampling is performed around the predicted centroids.

\subsection{Generative Model}\label{seg:generative}
Deep generative models are based on estimating the probability distribution of a dataset 
$X$
where the probability distribution 
$P_X$
is unknown. In particular, the assumption is made that the unknown distribution 
$P_X$ 
can be approximated by a parametric distribution $P_\omega$; where $\omega \in \Omega $ with $\Omega$ defined as a parametric space.  $P_\omega$ should be marginalized over a latent variable
$Z$
\begin{equation}
  p_\omega(\vector{x})=\int_Z p(\vector{x}|\vector{z})p(\vector{z})d\vector{z}
\end{equation}
There are mainly two well-known methodologies that utilize such approach for deep generative models, one is GAN \cite{10.5555/2969033.2969125}, where the density $p_\omega$ does not have an explicit analytical form. Instead, a random variable $Z$ is sampled from a specified random distribution, and a deterministic function $G_\omega: Z \rightarrow X$ aims to approximate the target distribution $P_X$. This is achieved by optimizing the Jensen-Shannon divergence \cite{e58ea65b62b04faf8b8373b494bcae5d}. The second approach, Variational Auto-Encoders (VAE) requires learning an approximation of the posterior distribution $P_{Z|X}$ and the marginal distribution $P_X$. The distribution $q_\omega (z|x)=\Normal(\mu_\omega(x),\Sigma_\omega(x))$ is used to approximate the conditional distribution with the Kullback–Leibler divergence loss between marginal distributions. While the VAE and GAN have been studied thoroughly in the literature, empirically GAN has been found to generate better quality results compared with the VAE's approach.

In order to improve the training stability of GAN models, a method was proposed in \cite{arjovsky2017wasserstein} which employs the Earth-Mover distance to measure the similarity between distributions, called WGAN. This approach offers the benefit of providing smooth measures even in scenarios where the distributions are completely overlapping or disjoint.  

Initially, in \cite{arjovsky2017wasserstein}, weight clipping was proposed as a method to stabilize the training of a WGAN model. However, choosing the right clipping parameters can be challenging, and setting them to values that are too large or too small can result in slow convergence or vanishing gradients. To address this issue, the authors in \cite{10.5555/3295222.3295327} introduced a gradient penalty approach that penalizes the model gradients if the Lipschitz constraint is violated. Specifically, if the critic function $f$ has a gradient norm greater than 1, a penalty term is added to the loss function to encourage the model to stay within the Lipschitz constraint as follows:

\begin{equation}
    \begin{split}
        L=
\underbrace{
    \ExpectedValue_{x \BehavesLike p_r(x)} [f_\omega(\vector{x})] - \ExpectedValue_{z\BehavesLike p_r(z)} [f_\omega(g_\rho (\vector{z}))]
}_{\text{Original critic loss}} \\
+
\underbrace{
\lambda \ExpectedValue_{\hat{x}\BehavesLike p_{\hat{x}}}
[
 (
\Norm{\Gradient_{\hat{x}}f_\omega(\vector{\hat{x}})}
  -1
 )^2
]
}_{\text{Gradient penalty}}
    \end{split}
    \label{eq:wgangp}
\end{equation}
where $p_r(x)$ and $p_r(z)$ represent respectively the distributions over the real multidimensional data $\vector{x}$ and the noise input vector $\vector{z}$. $\lambda$ is a penalty coefficient to weight the gradient penalty, $\vector{\hat{x}}$ sampled from the generator $g_{\rho},\rho \in \Omega$ and $\vector{x}$ within a $t$ uniformly sampled between $0$ and $1$.

However, from the Optimal Transport theory \cite{rout2022generative}; the Wasserstein distance is the transport ground cost $c(\vector{x},\vector{y})=\|\vector{x-y}\|$ and it can be shown that GANs are sensitive to hyper-parameters and difficult to train given that the transport map is discontinuous and DNNs can represent only continuous maps.  A proposed solution to this problem is to compute the optimal transport map between a continuous set and a latent distribution, as outlined in \cite{Gu2021}.

In our case, we propose to use a forward diffusion process as input to train the WGAN; defined as follows:
\begin{equation}\label{eq:forward}
    \vector{x_t}=\sqrt{\alpha_t}\vector{x}_{t-1}+\sqrt{(1-\alpha_t)}\vector{\epsilon}_{t-1}
\end{equation}
where $\vector{\epsilon}_{t-1},\vector{\epsilon}_{t-2},... \sim \Normal(\vector{0},\vector{I}),\alpha_t \in \Real$, this process creates a continuous map between the training input data and the latent distribution. Using latent distributions to train GANs is not a new concept, other works in the field of computer vision have used VAEs to represent the latent distribution of GANs with the objective of encoding the conditioning of the model; however this methodology requires training both VAE and WGAN at the same time, which could be challenging to find the perfect ratio between the approximation of the continuous latent function and the desired output. The SDE approach adds the benefit of establishing a continuous map from the input space to the latent space without the need to train the encoder and the GAN at the same time and without the search of the optimal radius between encoding and reconstruction. Also, our approach has the advantage that the latent variable \(x_t\) can be estimated in a single step using the reparameterization trick by a simple function in Eq. \eqref{eq:backward}, with $\hat{\beta}_t \in \Real$:

\begin{equation}\label{eq:backward}
    \vector{x_t} = \sqrt{\hat{\beta}_t} \vector{x}_0 + \sqrt{(1 - \hat{\beta})_t} \vector{\epsilon}_t
\end{equation}

This method does not require training and querying an additional DNN, which reduces sampling time. Additionally, this continuous function accounts for cases where previously unseen samples from the input space are covered by the path followed by a new input \(\vector{x}_0\) at time \(t\).

\subsection{Clustering}
We aim to simplify the training of the model given the number of data samples. Typically, other approaches that use DNNs involve tuples of image inputs and vector outputs to represent different samples from \(\StateSpace{C}_{free}\) used to train the approximation. However, approximating each image/sample of \(\StateSpace{C}_{free}\) in a large dataset requires processing a significant number of images per batch, which can slow down performance when the dataset is relatively large. 

In our approach, we propose to estimate the entire list of waypoints in a single step. However, this is impractical due to the infinite number of samples in a continuous \(\StateSpace{C}_{free}\). Instead, we aim to learn parameters that can generate an approximation of the total number of waypoints. One possible method is to use clustering techniques, which only require learning the parameters of the distribution to reconstruct the original space.

However, methods like Gaussian Mixture Models (GMMs) pose challenges during training. The random selection of initial centroids and the problem of solving the approximation using Expectation-Maximization often lead to different models even with the same parameters. This variability in cluster assignment means that the DNN may struggle to find a continuous map. In addition; when GMMs parameters are learned in the context of sampling based path planners; it can be challenging to condition the model \cite{teachingRobot} \cite{9712347}.

As an alternative, we propose using affinity propagation \cite{doi:10.1126/science.1136800}. Unlike GMMs, which require specifying the number of clusters, affinity propagation identifies exemplars directly from the dataset. It does not require defining the number of clusters in advance.

Affinity propagation operates by exchanging two types of messages between data points: the responsibility function, which measures how well a proposed data point represents its neighbors, and the availability function, which accumulates the responsibilities from other data points. This iterative process continues until convergence or a fixed number of iterations is reached. This method reduces the variance in cluster locations derived from the training dataset of waypoints in \(\StateSpace{C}_{free}\), as the exemplars are estimated using the entire dataset without relying on random sampling to form clusters. 
To approximate the set of waypoints, we use each exemplar as the mean of a Gaussian distribution and generate points around each exemplar with an initial standard deviation \(\sigma_0=0\). If we are unable to capture values in \(\StateSpace{C}_{free}\), we increase the standard deviation by following a vector of values \(\mathbf{\sigma} = \{\sigma_0, \dots, \sigma_k\}; k \in \NaturalNumbers; \sigma_k > \sigma_{k-1}; \sigma \in \Real^+\), instead of directly obtaining the sampled \(\StateSpace{C}_{free}\) configurations from \(f_\omega\).

\begin{figure}[h!]
\centering
  \begin{subfigure}[t]{0.49\columnwidth}
      \centering
	\def\svgwidth{\textwidth}
    \includegraphics[width=\linewidth]{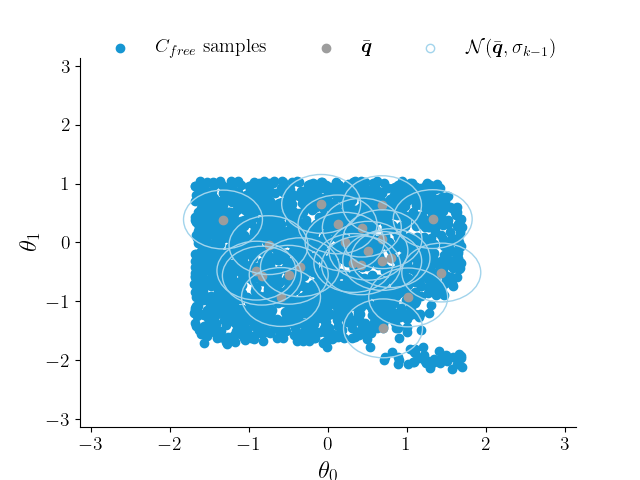}
    \caption{}

    \label{fig:originalCF}
\end{subfigure}
% \vfill
  \begin{subfigure}[t]{0.49\columnwidth}
      \centering
	\def\svgwidth{\textwidth}
    \includegraphics[width=\linewidth]{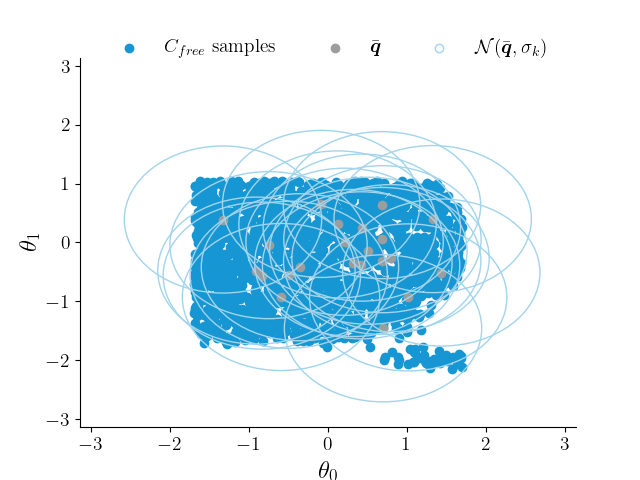}
    \caption{}

    \label{fig:originalCF1}
\end{subfigure}
  \caption{Projection of the first two joints of the Baxter manipulator's arm and its affinity points $\bar{\bm{q}}$. The circle represents the standard deviation $\sigma$. A fixed $\sigma$ could miss the original points around the exemplars, as shown in \figref{fig:originalCF}. However, if we increase $\sigma$, we eventually also increase the probability of sampling the missing \CF{}-configurations, as shown in \figref{fig:originalCF1}}
\label{fig:comparisonSampling}
\end{figure}

While the exemplars are not necessarily the means of Gaussian mixtures, we know in advance that their neighborhoods contain the original configurations in \(\StateSpace{C}_{free}\). Furthermore, there exists a \(\delta > 0\) such that the ball \(B_\delta(\bar{q})\), centered at the exemplar \(\vector{\bar{q}}\), includes its affinity points. The goal is to bias the sampler as close as possible to the original \(\StateSpace{C}_{free}\), thereby reducing the number of queries in \(\StateSpace{C}_{obs}\). This process is illustrated in \figref{fig:comparisonSampling}.

\subsection{Image Representation and Conditioning}\label{sec:ordering}
\begin{figure}[h!]
    \centering
 \includegraphics[width=0.5\linewidth]{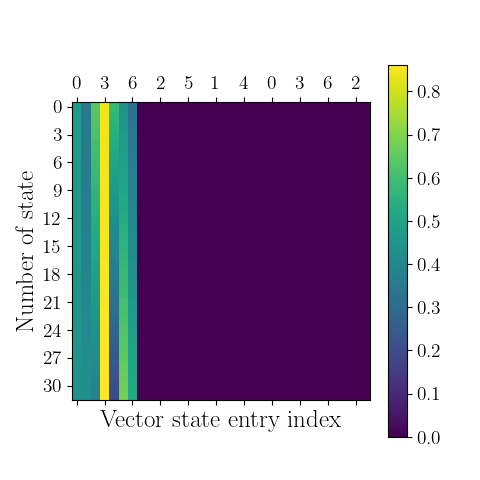}

    \caption{In the case of a constrained path, such as the shortest path, there is already an inherent ordering relationship between the configurations. In this example, we observe an almost continuous gradient between the different states in the matrix representation of a collision-free path, each entry of the waypoints are scaled.}
  \label{fig:orders0}
\end{figure}

Following the universal approximation theorem, DNNs are able to approximate any continuous function given an arbitrary number of activation functions. Thus, it is to our advantage to represent the training data problem as a continuous function. In our case, given that we have \cfree{} $\subseteq \mathbb{R}^n, n \geq 2$, a non-constrained approximation of \cfree{} represented as an image will rarely be continuous pixel-wise. However, the waypoints in \cfree{} that are part of a constrained path have a hierarchical ordering. Particularly, in the case of the shortest path; the affinity points of the set of waypoints can be represented an image with an almost continuous gradient; as shown in \figref{fig:orders0}.

\begin{figure}[h!]                                                                                             
\begin{center}                                                                                                 
  \begin{subfigure}{\columnwidth}                                                                              
                          \centering                                                                                     
 \includegraphics[width=\linewidth]{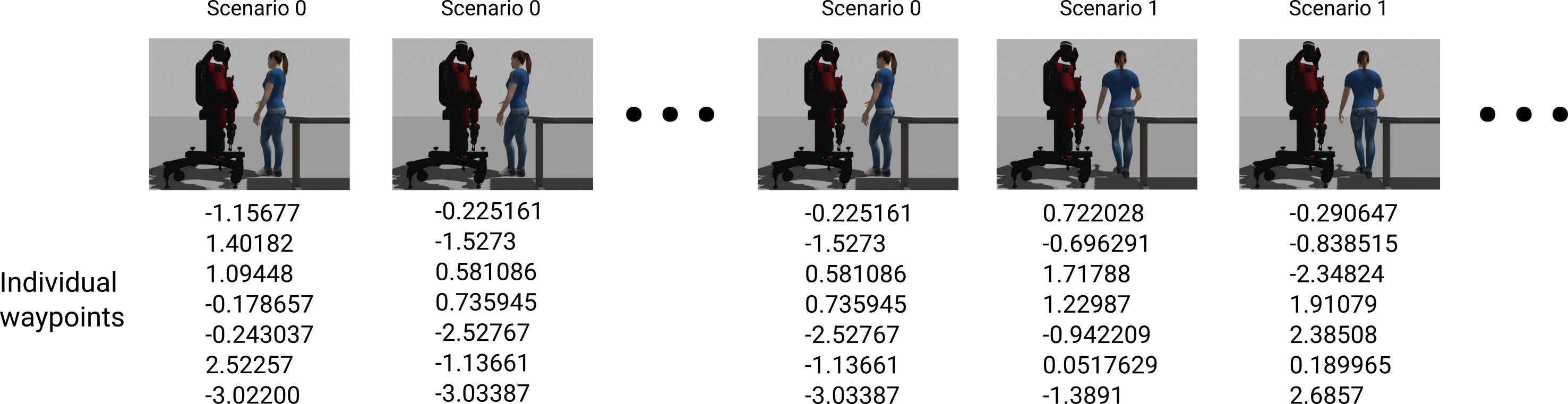}                                                              
    \caption{Conventional dataset for training \CF{} generative models}   
\label{fig:figureDataseta}
                                                                                                               
  \end{subfigure}                                                                                              
  \hfill                                                                                                        
  \begin{subfigure}{\columnwidth}   
      \centering
      \includegraphics[width=\linewidth]{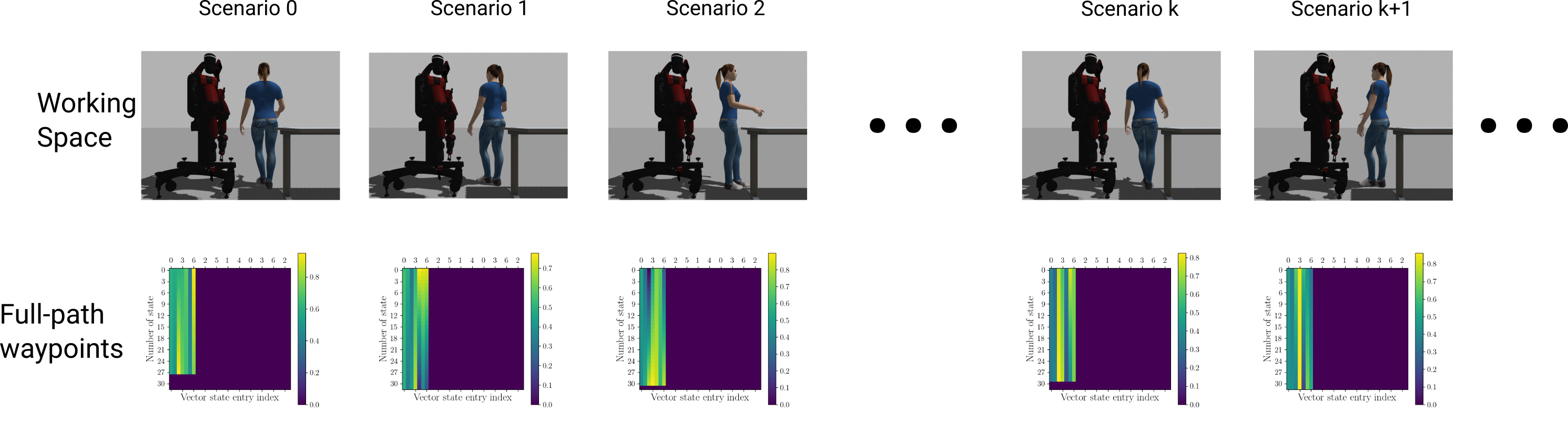}                                                            
    \caption{Proposed representation dataset for training waypoints in \CF{} generative models. The cardinality of the input working spaces is lower than the conventional representation}                                                                                           
\label{fig:figureDatasetb}

 \end{subfigure}                                                                                              
                                                                                                               
\end{center}                                                                                                   
  \caption{Conventional and proposed datasets for waypoints in \CF{} generative model training. In \figref{fig:figureDataseta}, we show how the dataset during training would repeat the same working space RGB-D for every waypoint sampled from \CF{}; thus requiring to increase the number of samples to be processed in the same epoch. In \figref{fig:figureDatasetb}, the representation only requires a paired matrix of centroids to represent all the waypoints, which decreases the total number of samples per epoch and decreases the time of learning. }                                               
  \label{fig:orders}    
\end{figure}

The matrix representation effectively reduces the paired training data between the RGB-D and each of the waypoints in \CF{}. Thus, we transform the problem to an image-to-image function, where each image represents the whole path in \CF{}. This process is exemplified in \figref{fig:orders}.

% \subsection{Image conditioning}\label{sec:imageconditioning}

Next, to overcome the challenges of the adversarial training of WGAN, we used the forward diffusion process on the input RGB-D images as explained previously.

Using a forward diffusion process as a latent variable for a GAN in the context of sampling-based planners is not a novel concept. In the work of \cite{pmlr-v164-ortiz-haro22a}, a WGAN is conditioned by the forward diffusion process of the RGB-D workspace (WS); however, the path conditioning can only be represented as a mask of the RGB-D, which limits the constraints that can be effectively captured. Similarly, the work of \cite{9636189} uses the forward diffusion process as a latent variable for the GAN's generator, but it requires a more structured approach to train the model in 3D workspaces. Other approaches necessitate discretizing the \CF{} of the robot \cite{pmlr-v164-lai22a}, which compromises the precision of the model.

To be able to represent diverse conditions outside of a mask of the original RGB-D; we also added an extra channel to represent the start state and the goal state of the path, which we refer to as RGB-D+. Given that the start and goal states can be represented by a vector of 14 entries, we propose to fill the rest of the condition channel with the positional encoding proposed in \cite{NIPS2017_3f5ee243}, which has been shown to improve gradient propagation \cite{10.5555/3305381.3305510}. This representation gives us the chance of not only represent information encoded as an image in the WS like the one presented in \cite{pmlr-v164-ortiz-haro22a}; any other numerical information can be represented as such given that it can be fit in the same dimension as the resolution of the input image.

\begin{figure}
  \centering

     \includegraphics[width=0.6\linewidth]{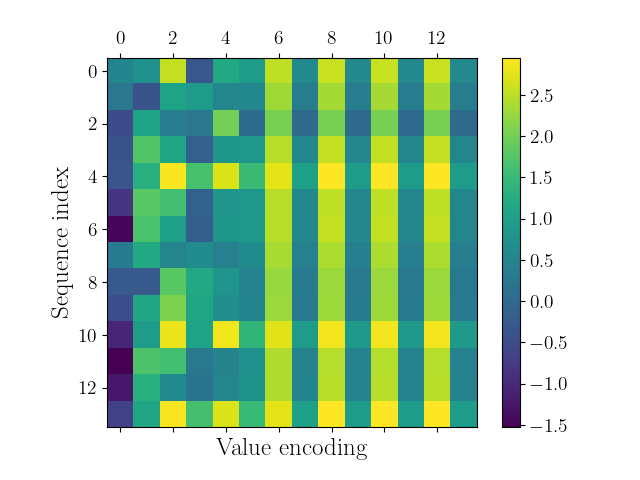}
  \caption{Extra channel to add conditioning to the generator. For instance, when training the model to generate waypoints for the shortest path, we include the start and end states of the path in this extra channel; the condition consists of 14 entries, where 7 corresponding to the start state and 7 to the end state. Each row represents different encoding.}
\label{fig:positionalEncoding}
 
\end{figure}

Our proposed representation helps us to encode the information directly into the image input $\vector{y}_0$, which would be used as the latent for $\vector{y}_t$. This change will help reduce the complexity of the DNN, given that the condition is already inside the RGB-D+, and we can exploit image processing techniques like convolutions to avoid the need for fully connected layers between all entries of the condition's vector. An example of the condition channel is presented in \figref{fig:positionalEncoding}.

\subsection{Planner}
To guide the path towards the desired region in \CF{}, we utilize a technique inspired by \cite{Wang2020NeuralRL}; where a 2D \CS{} is represented as an image for the DNN to be learned by convolution layers. Our approach involves using a DNN's model generator trained by an adversarial loss function as a sampler for the RRT path planning algorithm. However, we made a modification in our implementation to reject the sampling from the training model given the probability of the data around an estimated waypoint being in the complement of \CF{}.

\begin{algorithm}
    \caption{RRT with WGAN-GP cluster bias}
    \label{alg:biasedRRT}
    \KwData{
      $\vector{q}_{init},
      \vector{q}_{goal},
      \vector{\epsilon}=[\vector{\epsilon}_0,\vector{\epsilon}_1,...,\vector{\epsilon}_t], t\in \NaturalNumbers, 
      \vector{\epsilon}_j \in \Normal(\vector{0},\vector{I})$ ,
      $\vector{y_0} \in \text{RGB-D+}: [0,1]^{5\times 32\times 32},$
      $\vector{\sigma}=[\sigma_0,\sigma_1,...,\sigma_k], k \in \NaturalNumbers, 
      \sigma_i \in \Real^+$ and $\sigma_i>\sigma_{i-1}$ is an increasing amount of perturbation for the exemplars $\vector{\bar{q}}$, generator $f_\omega(\cdot)$. 
      $m$ is the number of samples taken from each
      $\Normal(\vector{\bar{q}},\sigma_{i}\vector{I})$
    }
    \KwResult{
        $
        G
        $
    }
    $s\leftarrow 0$\;
    $V\leftarrow \vector{q}_{init},E\leftarrow \emptyset$\;
    $\vector{y}_t \leftarrow \sqrt{\hat{\beta}_t}\vector{y}_0+\sqrt{(1-\hat{\beta}_t)}\vector{\epsilon}_t$\;

    $\vector{\bar{q}} \leftarrow f_\omega(\vector{y}_t)$\;
    % $BiasSampler=\cup^{k}_{l=0} \cup^{m}_{j=0}\Normal(\vector{\bar{q}},\sigma_l)$\;
    % $Sampler\leftarrow BiasSampler$\;
    \For{$h=1,...,l$}{
        \If{$h$ $mod$ $m ==0$}{
            $s \leftarrow s+1$
        }

        \If{$s<= k$}{ 
            \If{$\Normal(\vector{\bar{q}},\sigma_{s}\vector{I})\cap$\CF{}$\neq \emptyset$}{
                $\vector{q}_{rand}\leftarrow SampleFree(\Normal(\vector{\bar{q}},\sigma_{s}\vector{I}))$\;

            % $Sampler\leftarrow \Normal(\vector{\bar{q}},\sigma_{s}I)$
            }
            \Else{
                $\vector{q}_{rand}\leftarrow SampleFree(\mathcal{U}($\CS{}$))$\;
            }
        }
        \Else{

            $\vector{q}_{rand}\leftarrow SampleFree(\mathcal{U}($\CS{}$))$\;
}
        $\vector{q}_{nearest}\leftarrow Nearest(G=(V,E),\vector{q}_{rand})$\;
        $\vector{q}_{new}\leftarrow Steer(\vector{q}_{nearest},\vector{q}_{rand})$\;
        \If{$ObstacleFree(\vector{q}_{nearest},\vector{q}_{new})$}{
            $V\leftarrow V \cup \{\vector{q}_{new}\}$\;
            $E\leftarrow E \cup \{(\vector{q}_{nearest},\vector{q}_{new})\}$\;
        }
        \If{$\vector{q}_{new}==\vector{q}_{goal}$}{
          \Return $G=(V,E)$\;
        }

    }
    \Return $G=(V,E)$\;
\end{algorithm}
% \begin{table*}
%  \centering 
%   \begin{tabular}{c|c|c|c}
%     Algorithm & Max Planning Time (s) & Success \% & Average planning time (s)\\
%     \hline
%     RRT & 0.1 & 18\% & 0.07 $\pm$ 0.02 \\
%     RRT & 0.2 & 35\% &  0.10 $\pm$ 0.05\\
%     RRT & 0.5 & 56\% &  0.19 $\pm$ 0.12\\
%     RRT-WGAN & 0.1 & 33\% &  0.05 $\pm$ 0.01 \\
%     RRT-WGAN & 0.2 & 40\% &  0.08 $\pm$ 0.05 \\
%     RRT-WGAN & 0.5 & 55\% &  0.16 $\pm$ 0.18\\
%   \end{tabular}
%   % }
%   \caption{Results of planning using an unconstrained \CF{}.}
%   \label{tab:unconstrained}
% \end{table*}

One advantage of our proposal compared with other approaches in the field of path planning using random trees and DNNs is that we can estimate when our approximation has most likely failed. Given that the probability of selecting a sample on the tail of the Gaussian distributions of each exemplar decays exponentially, we have an upper bound to our trained model to know when the generator $f_\omega$ most likely failed to approximate a bounded \CF{}. 

Given that we only have an approximation of the affinity points and lack
information about their neighboring waypoints, we aim to recover some of the
missing waypoints by sampling around $\vector{\bar{q}}$. Although we cannot
guarantee that the process will recover all the original waypoints by sampling
from $\Normal(\vector{\bar{q}}, \sigma_i\vector{I})$, we increase the value of $\sigma_i$ with the goal of identifying as many missing neighboring waypoints as possible within an unknown neighborhood.

We know that $P(|\vector{q} - \vector{\bar{q}}_i| > \sigma) \sim 0.3173$
 and that sampling from
$\Normal(\vector{\bar{q}}, \sigma_i\vector{I}) \cap$\CF{}$\neq \emptyset$ should hold for sufficiently large $m$, provided that the affinity points were well-approximated during training. If this condition does not hold, it is likely that the trained model failed to capture the true waypoints for $\sigma_i$ with sufficiently large $m$.

% Given that we only have some approximation of the affinity points and we do miss some information about its neighbors waypoints, we are expecting to get some of the missing waypoints by sampling around $\bar{q}$. While we do not know if the process would be able to recover all the original waypoints by sampling from $\Normal(\bar{q},\sigma_i)$; we increase the value of $\sigma_i$ in hopes to find as many of the missing waypoints neighbors in an unknown neighborhood; however, we know that $P(|q-\bar{q_i}|>\sigma) \sim 0.3173$ and that sampling from $\Normal(\bar{q},\sigma_i)\cap$\CF{} $\neq\emptyset$ for $m$ samples, with $m$ big enough should be true in the case that a good approximation of the affinity points was learnt, if that is not the case, it is very likely that the trained model has failed to capture true waypoints. Thus, we can detect if the model has likely failed to get good approximations of the learnt affinity points by rejecting sampling from the learnt distribution if the condition $\Normal(\bar{q},\sigma_i) \cap$ \CF{} $= \emptyset$ is met for each $\sigma_i$ with $m$ big enough.

% By increasing $\sigma_i$ until we reach the bound of the configuration space we can test if the learnt model is not good enough to be able to generate points in \CF{}, if that is not the case we switch to sampling from the uniform distribution. With a defined bound for the bias sampling induced by $\sigma_0,...,\sigma_k$, the mixture model sampler for RRT could be defined as follows in equation \ref{eq:completeness} with $m$ samples per $\Normal(\bar{q},\sigma_i)$ as:
By increasing $\sigma_i$ until the boundary of the configuration space is
reached, we can evaluate whether the learned model is inadequate for generating
points in \CF{}. If this condition is met, we switch to sampling from the
uniform distribution. This proves that our algorithm is probabilistic complete
following \cite{770022}, when the total number of samples $l \in \NaturalNumbers$ goes to infinity. The proposed algorithm is described in Algorithm
\ref{alg:biasedRRT}. Where $\mathcal{U}$ is the uniform distribution, and
$SampleFree(\cdot)$ is a function that samples the input distribution until it finds a configuration in \CF{}. 

%Given a defined bound for the bias introduced by $\sigma_0, \dots, \sigma_k$, the mixture model sampler for RRT can be defined as shown in Equation \ref{eq:completeness}, using $m$ samples per $\Normal(\bar{q}, \sigma_i)$ as follows:
%
%\begin{equation}
%  q_h\sim
%\begin{cases}
%  h > m*k &\mathcal{U}(C)\\
%    h \leq m*k & \cup^{k}_{i=0} \Normal(\bar{q},\sigma_i)
%\end{cases}
%    \label{eq:completeness}
%\end{equation}
%where $h,m \in \NaturalNumbers$; when $h \rightarrow \infty$, $q_{h}\sim\mathcal{U}(C_{free})$ with $\mathcal{U}$ as the uniform distribution;  which proves that our algorithm is probabilistic complete following \cite{770022} with $h>k*m$. The proposed algorithm is described in Algorithm \ref{alg:biasedRRT}.  

\section{Experimental results}
\begin{table*}
 \centering 
  \begin{tabular}{c|c|c|c|c}
    Algorithm & Max Planning Time (s) & Percentage of success & Average planning time (s) & Average Length rad\\
    \hline
    RRT & 0.1 & 35\% & 0.07 $\pm$ 0.02 &10.12 $\pm$ 3.18\\
    RRT* & 0.1 & 30\% &  0.1 $\pm$ 0.0&10.04 $\pm$ 3.11\\
      RRT-WGAN & 0.1 & \highlighti{red}{ 63\% } & \highlighti{red}{ 0.06 $\pm$ 0.01} & \highlighti{red}{9.91 $\pm$ 2.86}\\
          \hline\hline
    RRT & 0.2 & 60\% &  0.11 $\pm$ 0.04&10.65 $\pm$ 3.25\\
    RRT* & 0.2 & 58\% &  0.2 $\pm$ 0.0&10.57 $\pm$ 3.20\\
      RRT-WGAN & 0.2 & \highlighti{red}{75\% } &  \highlighti{red}{0.08 $\pm$ 0.04} &\highlighti{red}{10.4 $\pm$ 3.15}\\
          \hline\hline
      RRT & 0.5 &\highlighti{red}{ 94\% } &  0.15 $\pm$ 0.1&10.91 $\pm$ 3.33\\
      RRT* & 0.5 & 87\% &  0.5 $\pm$ 0.0&\highlighti{red}{10.73 $\pm$ 3.15}\\
    RRT-WGAN & 0.5 & 87\% &  \highlighti{red}{0.11 $\pm$ 0.1}&10.81 $\pm$ 3.32\\
  \end{tabular}
    \caption{Results of constrained planning of a collision-free path using a Baxter robot and a human inside its working space in previously unseen scenarios. The constraint is finding the shortest path. The red rectangles represent the best result in each metric given different maximum planning times.}
  \label{tab:constrained}
\end{table*}

All the models are trained on a system with 2 x Intel Gold 6148 Skylake, 16 GB of RAM and 2 x Nvidia V100SXM2. For deployment, we use Ubuntu 20.04 running on a 3.60 GHz × 8 Intel Core i9-9900KF processor, 16GB RAM on Nvidia RTX 2070.

We propose a set of experiments to examine the effectiveness of learning the waypoints of a constrained path. These waypoints are used to bias the sampling of \cfree\, states to find a collision-free path, as described in Section \ref{sec:methodology}. The constrained approximation illustrates how well the conditioning of \CF{}\, is encoded in the RGB-D+ input and also provides insight into how the implementation might work in real-world problems.

We use a combination of 100 WSs/\CF{}s with 100 different starting and goal configurations. All WSs and waypoints are derived from a simulated Baxter's 7-DOF right arm. All input images of the scenarios were resized to 32 x 32 pixels. We utilized the PyTorch Lightning framework with Adam optimizer parameters derived from \cite{10.5555/3295222.3295327} for training. Our experimental hyperparameters consisted of a learning rate of $4 \mathrm{e}{-5}$, a batch size of 128, a regularization coefficient $\lambda$ of 10 for the gradient penalty, and $n_{\text{critic}}=5$ training iterations per generator iteration. All \CS{}s are scaled between [0,1] in their matrix encoding. Our code is available on Bitbucket\footnote{\url{https://bitbucket.org/joro3001/imagewgangpplanning/src/master/}}.

We employ a simulated RealSense RGB-D sensor to represent the conditioning factor in our experiment. This image captures the obstacles' representation within the robot's operational space. In cases of constrained paths by path length, we add the initial and final states of the shortest path as an extra channel. To streamline the experimentation process, we opted for a human model spawning in a random position and orientation within Baxter's working space. The initial and final configurations of the arm are chosen randomly.

For the test, we train the generator using waypoints of the shortest path between random goal and start configurations. The shortest path depends on the initial and goal configurations; thus, the goal and start states condition the generation of the waypoints to generate the RGB-D+ training data.  The original waypoints of the shortest path were obtained by running RRT* for 30 seconds. We used RRT* paths as training data, which provided a more stable distribution that did not fluctuate significantly when the configuration space was changed. We used path length as the minimization objective for the generator training, as it provides a reliable measure of the quality of the generated paths.

To implement the trained sampler, we utilized the Open Motion Planning Library (OMPL) \cite{kingston2019exploring} implementations of RRT and RRT*. To compare how well the constraints were maintained by the trained model, we ran 3500 different, previously unseen scenarios with random goals and states, similar to the previous section. Each algorithm was run for 0.1, 0.2, and 0.5 seconds. The results are presented in Table \ref{tab:constrained}.

As shown in Table \ref{tab:constrained}, our algorithm improves the success rate, planning time, and average length compared to the implementations of RRT and RRT* when the time constraint is between 0.1 and 0.2 seconds. When the planning time increases, RRT is able to improve its success rate because the uniform sampler starts the process earlier when the scenario is more challenging for the trained generator and too far of a good approximation of the original waypoints. Without timing constraints, RRT* begins to converge to the optimal path cost; however, in scenarios where query time is critical, our approach generates a collision-free path that improves the cost compared to RRT* at query times similar to RRT, as seen in the 0.5 seconds case. Additionally, for cost constrained paths, we see that our algorithm increases the success rate of the planning task, as it spends less time reaching the boundary set by $\sigma_k$, given that most of the time constrained waypoints is a subset of \CF{}. This clearly demonstrates that our approach effectively increases the success rate, reflecting the probabilistic completeness of our algorithm given that the success rate increases as the maximum planning time increases.

%What did we learn?
%What conditions does it work...
\section{Conclusion and Future Work}
%What did we do? What do we don't know?

In this work, we have presented a novel approach for training WGAN-GP models conditioned by continuous latent matrices, which can be utilized for tasks related to waypoints prediction and path planning. We also proposed using the parametric learned model to evaluate whether the approximation has failed in predicting the waypoints in \CF{}. Additionally, we explored incorporating extra channels from the RGB-D working space to constrain the path to a specific cost function.

The experiments indicated that using images as a representation of waypoints in \CF{} stabilizes training and simplifies the deep neural network (DNN) models by utilizing convolutional networks instead of fully connected DNNs. The inclusion of high-dimensional ordering contributes to creating almost continuous training data in the image space. However, it is important to note that poor approximations of the waypoints could increase the planning time to be in par with the original sampler of the path planner, as regions not part of the collision-free path will be explored first.

The results of our experiments demonstrate that our proposed model is capable of generating collision-free paths in unknown scenarios with an improved success rate and reduced running time compared to conventional path planning algorithms such as RRT and RRT*. This is particularly useful when these algorithms are constrained by a specific running time, making our approach valuable for real-world scenarios.

In the case of having a bad approximation of the waypoints, our proposed approach is capable of rejecting such samples and revert to a uniform sampler without input from the user about the ratio between samplers, a capability non-previously seen in other works.

We have also shown the effectiveness of our method in planning paths in a 7-dimensional space for a humanoid-manipulator robot. To establish the broader applicability of our method, it is necessary to extend it to higher-dimensional spaces and non-stationary robots. This is critical for demonstrating its usefulness in solving diverse problems in real-world applications and scenarios. Future work will focus on exploring the feasibility of this extension and evaluating the performance of our method on more complex tasks and scenarios.

The convergence of affinity propagation could potentially generate more exemplars than the matrix can accommodate. Therefore, further research is needed to explore methods for reducing the amount of data required to represent the waypoints in \CF{} as a matrix representation.

\section{Acknowledgement}
We acknowledge the computing resources provided by Calcul Qu\'ebec and the Digital Research Alliance of Canada.
\bibliographystyle{IEEEtran}

\bibliography{IEEEabrv,refs} % Entries are in the refs.bib file

\end{document}